\documentclass[final]{cvpr}

\usepackage{times}
\usepackage{epsfig}
\usepackage{graphicx}
\usepackage{amsmath}
\usepackage{amssymb}

\usepackage{tabularx}
\usepackage{enumerate}
\usepackage{multirow}
\usepackage{booktabs}
\usepackage{amssymb}
\usepackage{pifont}
\usepackage{setspace}

\usepackage{makecell}
\usepackage{nopageno}

\usepackage[pagebackref=true,breaklinks=true,colorlinks,bookmarks=false]{hyperref}

\def\cvprPaperID{5030} % *** Enter the CVPR Paper ID here
\def\confYear{CVPR 2021}

\begin{document}

\title{IQDet: Instance-wise Quality Distribution Sampling for Object Detection}

\author{{Yuchen Ma}, \ 
        {Songtao Liu}, \ 
        {Zeming Li}, \
        {Jian Sun} \        
        \\
        
        Megvii Technology \\
       \small{\{mayuchen, liusongtao, lizeming, sunjian\}@megvii.com}}

\maketitle

% #################################################################################
%                              abstract
% #################################################################################
\begin{abstract}
We propose a dense object detector with an instance-wise sampling strategy, named IQDet. Instead of using human prior sampling strategies, we first extract the regional feature of each ground-truth to estimate the instance-wise quality distribution. According to a mixture model in spatial dimensions, the distribution is more noise-robust and adapted to the semantic pattern of each instance. Based on the distribution, we propose a quality sampling strategy, which automatically selects training samples in a probabilistic manner and trains with more high-quality samples. Extensive experiments on MS COCO show that our method steadily improves baseline by nearly 2.4 AP without bells and whistles. Moreover, our best model achieves 51.6 AP, outperforming all existing state-of-the-art one-stage detectors and it is completely cost-free in inference time.

\end{abstract}

% #################################################################################
%                              Introduction
% #################################################################################

\section{Introduction}
\label{sec:intro}

\begin{figure}[t]
\centering
\includegraphics[height=0.6\linewidth, width=1.0\linewidth]{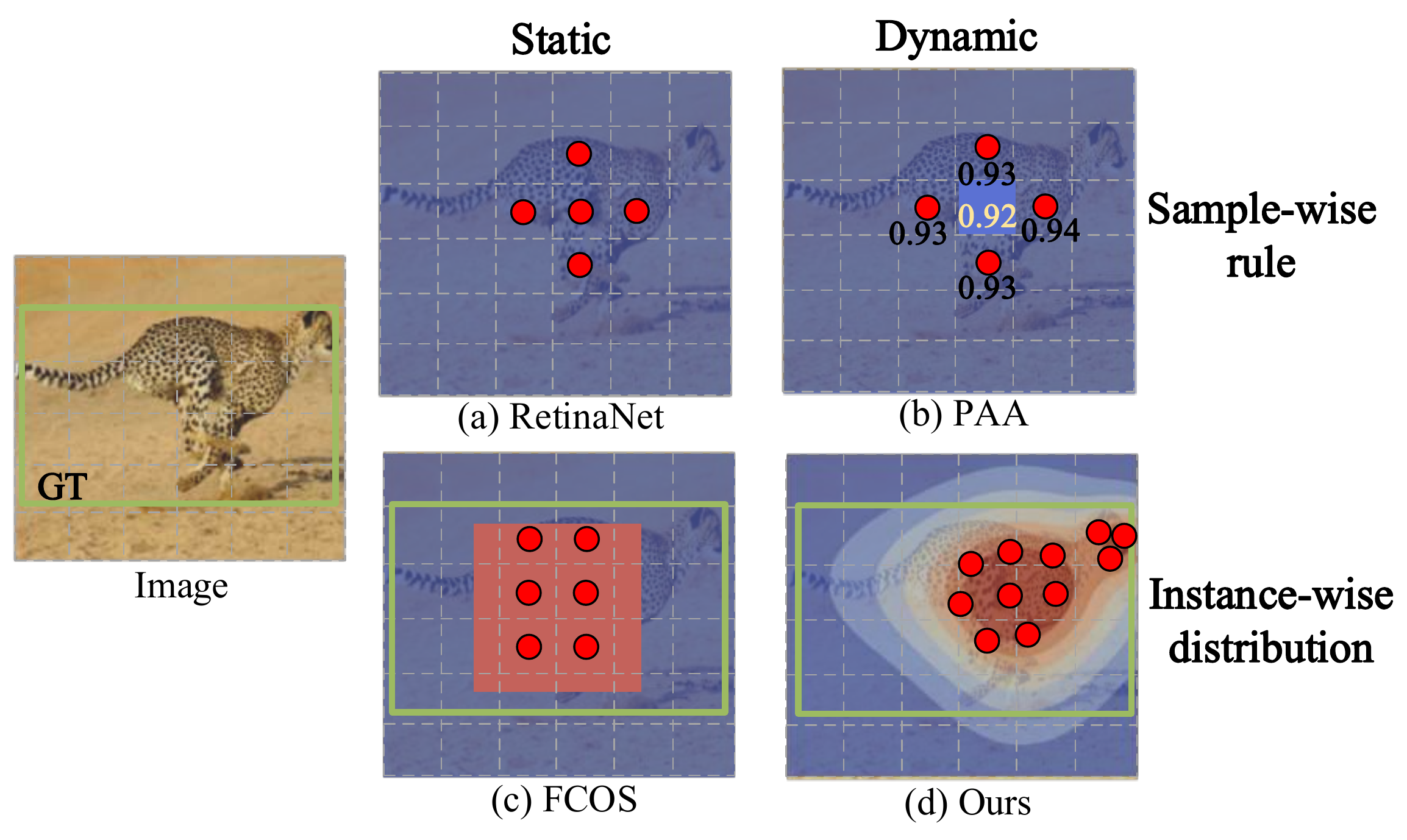}
\caption{Different sampling strategy of dense object detection. The green squares are GT and the red circles represent the positive samples. (a) \textit{Static} sampling strategies mainly base on the anchors' locations, while the \textit{Dynamic} sampling strategies are prediction-aware. (b) The IoUs of the predicted boxes are quite similar, but the predicted box with IoU=0.92 is compulsively assigned as a negative sample due to the slightly lower IoU, bringing inconsistent noisy samples for the training process. (c) FCOS uses a static center distribution for sampling. (d) IQDet samples training set according to an instance-wise dynamic distribution.}
\label{fig:abstarct}
\end{figure}

% For most CNN-based detectors, a dominant paradigm of representing instances of various shapes and sizes is to enumerate anchor boxes of multiple scales and ratios at every spatial location. Therefore, the selection and assignment of samples are critical to the performance of an object detector. 

Recent object detection methods have proposed various sampling strategies, assigning the predicted boxes to the ground truth according to their quality. These sampling and assignment strategies have led to great progress in modern object detection task.

As shown in Fig.~\ref{fig:abstarct}, the improvements in sampling strategies can be divided into two tendencies. (1) From \textit{Static} to \textit{Dynamic}: \textit{Static rules} like RetinaNet~\cite{focal_loss} assign anchor boxes to the ground truth according to the predefined quality of anchors (\emph{i.e.}, IoU), while \textit{Dynamic rules} (\emph{e.g.}, loss-based sampling in PAA~\cite{paa}) further boost the performance of the dense object detection, because they are prediction-aware and directly determined by the prediction quality. (2) From \textit{Sample-wise} to \textit{Instance-wise}: Assigned by the \textit{Sample-wise} rule like RetinaNet\cite{focal_loss}, a predefined anchor is assigned as a positive sample when its IoU is higher than the threshold. Meanwhile, based on \textit{Instance-wise} distribution, \cite{fcos, guided_anchor} propose a center-based distribution according to the ground-truth(GT)'s location, where the samples located inside the central region are assigned as positive samples. 
%These anchor-free methods without predefined anchor show more potential in terms of generalization ability. 
These two tendencies achieve better performance and gradually become the de-facto standard in object detection.

However, these sampling strategies might have a few limitations:
\label{sec:motivation}
(1) \textit{Static rules} are not learnable and prediction-aware (e.g. center region and anchor-based), which may be not always the best choice for some eccentric object. (2) Some \textit{Dynamic rules} like PAA might suffer from the noisy samples and per-sample quality rules, without jointly formulating a quality distribution in spatial dimensions, as shown in Fig.~\ref{fig:abstarct}(b). (3) They sample uniformly over regular grids of image owing to the dense prediction paradigm, which is difficult to assemble enough high-quality and diverse samples. These methods might either \textit{Static} or \textit{Sample-wise}, failing to completely solve these limitations. 
%Thus, to overcome these drawbacks above, the key components of training strategies are modeling a robust instance-wise quality distribution, which is dynamic and learnable.

In this paper, we propose an instance-wise(GT-wise) quality distribution for sampling to address these issues: instead of assigning each sample independently, the instance-wise sampling strategy selects training samples based on the quality distribution. To achieve dynamic sampling, we jointly model an instance-wise distribution from the network prediction, making it learnable and prediction-aware. Specially, we optimize the quality distribution encoding function via a variational encoder, approximating the semantic pattern and appearance of the instance. This dynamic quality distribution filters out noisy samples(shown in Fig.~\ref{fig:abstarct}(b)) and is easier to be learned than previous sampling strategies. Owing to the previous per-pixel assignment depends on IoUs of each predicted boxes, the assignment of PAA~\cite{paa} is unstable. Our assignment strategy is more robust and extracts the overall instance-wise feature to generate the assignment. And it can sample more high-quality samples, especially the location in the center of the Gaussian.

Further, we use this distribution to guide sample selection in both spatial and scale dimensions, as shown in Fig.~\ref{fig:abstarct}(d). Proper samples are resampled with higher probabilities and assigned by higher confidences. Meanwhile, different from the traditional per-grid sampling strategy, we sample more positives from high-quality and diverse predictions with a floating-number coordinate, which may avoid the overfitting problem and improve the performance efficiency.
The entire process of instance-wise quality assignment and sampling are learnable and thus can be easily optimized by back-propagation.

The contributions of this work are summarized as:
\begin{enumerate}[1]
\item Our main contribution is to propose an instance-wise quality distribution, which is extracted from the regional feature of the ground-truth to approximate each prediction's quality. It guides noise-robustly sampling and it is a prediction-aware strategy. 
\item Besides, we formulate an assignment and resampling strategy according to the distribution. It is adapted to the semantic pattern and scale of each instance and simultaneously training with sufficient and high-quality samples.
\item We achieve state-of-the-art results on COCO dataset without bells and whistles. Our method leads to 2.8 AP improvements from 38.7 AP to 41.1 AP on single-stage method FCOS. ResNext-101-DCN based IQDet yields 51.6 AP, achieving state-of-the-art performance without introducing any additional overhead.
\end{enumerate}
% \vspace{-3pt}

% #################################################################################
%                              Related Works
% #################################################################################
\begin{figure*}[t]
\centering
\includegraphics[height=0.46\linewidth, width=0.98\linewidth]{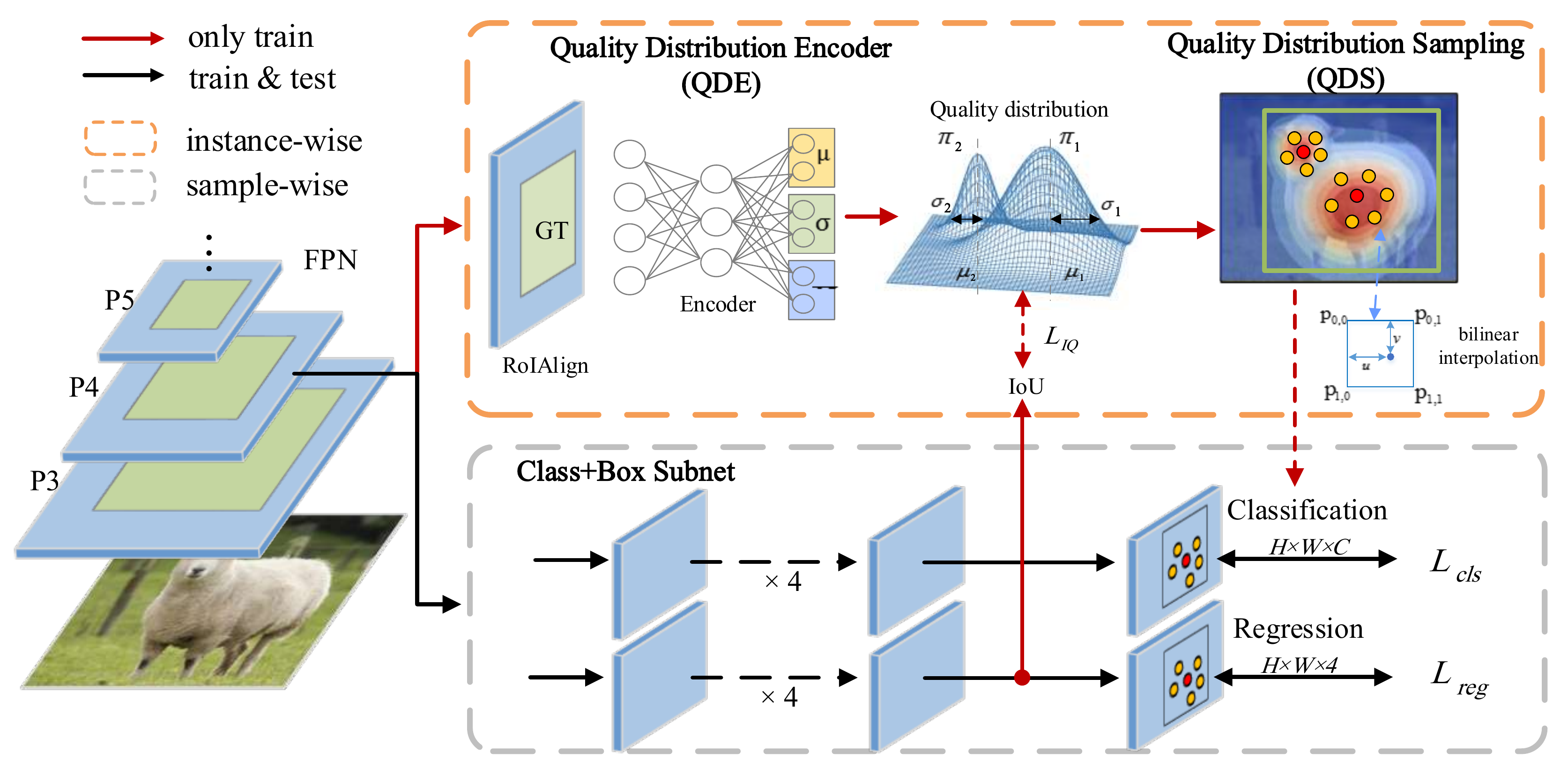}
\caption{The architecture of IQDet, where the parameters of the quality distribution ($\mu$,$\sigma$,$\pi$) are predicted by Quality Distribution Encoder (QDE), which represent location, scale, and mixing coefficient of each component respectively. The Class+Box Subnet uses the intermediate output (P3 - P7) from each feature-map of the feature-pyramid, which is similar to FCOS~\cite{fcos}. The subnets in the orange box are the instance-wise operations and the subnets in the gray box are the sample-wise operations. Our sampling strategy is completely cost-free in inference time, as the auxiliary structures only exist during training.}
\label{fig:architecture}
\end{figure*}

\section{Related Work}
\textbf{Label Assignment in Object Detection.}
The task of selecting which anchors are to be assigned as positive or negative samples has recently been recognized as a crucial factor that greatly affects the detector's performance~\cite{focal_loss, fcos, paa, borderdet, song2020fine}. CenterNet~\cite{objects_as_point} and FoveaBox~\cite{foveabox}, they both use center-sampling strategy to select the positive samples. GuidedAnchoring~\cite{guided_anchor} leverages semantic features to guide the anchor settings and dynamically mitigate the feature inconsistency with a feature adaption module. FreeAnchor~\cite{freeanchor} uses a bag of \textit{topK} anchor candidates based on IoU for every object and determines positive anchors based on the estimated likelihood. ATSS~\cite{atss} suggests an adaptive training sample assignment mechanism by the dynamic IoU threshold according to the statistical characteristics of instances. PAA~\cite{paa, LLA} adaptively separates anchors into positive and negative samples according to the detector’s learning status in a probabilistic manner, and they propose to predict the Intersection-over-Unions of detected boxes as a measure of localization quality. AutoAssign~\cite{autoassign} uses a fully differentiable label assignment strategy to automatically determines positive/negative samples by generating positive and negative weight maps. MDOD~\cite{mdod} proposed a network by estimating the probability density of bounding boxes in an input image using a mixture model and they got rid of the cumbersome processes of matching between ground truth boxes. POTO~\cite{end2end} uses prediction-aware label assignment to enable end-to-end detection.
These methods are either \textit{static} or \textit{sample-wise}, which might have a few limitations. We propose a quality distribution for instance-wise sampling strategy, which aims at solving these limitations and boosting the performance of dense object detection.

% \textbf{Feature Representation Learning.}
% There are several works that focus on the feature representation encoding to boost training~\cite{tang2014feature, zhuang2015supervised, bengio2013representation, autoencoders}. VAE~\cite{vae} proposed a formulation to learn with directed probabilistic models and used a variational approach for latent representation learning. It is useful for data representation learning and feature denoising. \cite{regularizing} uses an AutoEncoder to model the labels of semantic segmentation and encode the label structure of the instance. \cite{hao2020labelenc} introduce a novel label encoding function, mapping the ground-truth labels into latent embedding. They optimize the label encoding function via an AutoEncoder defined in the label space, approximating the “desired” intermediate representations for the target object detector. 

\textbf{Sampling Strategies in Object Detection.}
The most widely adopted sampling scheme in object detection is the random sampling from all candidates. OHEM ~\cite{ohem} mines hard examples which have larger losses at each iteration on the fly. 
% Libra R-CNN \cite{librarcnn} proposes IoU-balanced sampling as a scheme of hard negative mining, which samples an equal number of negative samples in different IoU ranges. 
% PISA \cite{pisa} focuses on prime samples than hard samples when training a detector, which uses IoU-HLR to rank the importance of samples to reweight the sampling. 
Cascade R-CNN ~\cite{cascadercnn} proposes that they only sample high-quality prediction to train, the detector is easier to overfitting, due to exponentially vanishing positive samples, which might be harmful to the performance of the detector. And it was shown to avoid the problems of overfitting at training and quality mismatch at inference. SAPD~\cite{sapd} jointly samples the anchor point as a group both within and across feature pyramid levels and soft-weighted anchor points and soft-sampled pyramid levels across all the pyramid levels. 
% Dynamic R-CNN\cite{dynamicrcnn} adopts the label assignment criteria (IoU threshold) and makes better use of the training samples and pushes the detector to fit more high-quality samples. 

% #################################################################################
%                              Methods
% #################################################################################
% \vspace{-3pt}
\section{Methods}
% \vspace{-3pt}
In this section, we first formulate a quality distribution as an instance-wise representation encoding, which is adapted to the semantic pattern of each instance and noise-robust. Next, based on this distribution, we show how to select samples in a probabilistic manner and train with more high-quality samples (Sec.~\ref{sec:quality_sampling}), which improves the overall performance by a large margin.

\subsection{Formulation of Quality Distribution Encoder}
\label{sec:quality_distribution_encoder}

As mentioned in ~\ref{sec:motivation}, previous sampling strategies~\cite{fcos, guided_anchor, atss, paa} suffer from these limitations above. Thus, the key idea is to introduce a robust instance-wise distribution, which is prediction-aware and is adapted to the semantic pattern of an object. 

We propose a novel distribution learning subnet, called Quality Distribution Encoder(QDE). The architecture of the encoder is demonstrated in Fig.~\ref{fig:architecture}. It is worth noting that all operators in Quality Distribution Subnet in the orange box are instance-wise. To effectively encode the instance-wise feature, we first extract the feature of an object according to the GT location and it is realized by applying the RoIAlign layer ~\cite{mask_rcnn} to each pyramid feature, where the input RoI is the ground-truth box. Specifically, the motivation of using GT feature is that extracting the regional feature of GT is properly aligning with the distribution assignment in spatial dimensions. Thus the encoder is an instance-wise extractor and estimates an overall quality distribution for an instance per pyramid level. To map the GT feature into an instance-wise representation, the encoder should capture the principal component of the ground-truth feature and then estimate an instance-wise quality distribution. Owing to the unknown underlying distribution is not easy to learn, a basic idea is to use an encoder to map the unknown distribution (e.g. quality distribution of per-sample) to a specific distribution like Gaussian. 

We choose Gaussian Mixture Model (GMM) to model the distribution. GMM enjoys the several characteristics mentioned in~\cite{gmm}: (1) It can form smooth approximations to arbitrarily shaped distribution. (2) The individual component may model some underlying set of hidden classes. Thus, according to the GMM modeling, we can estimate a distribution to approximate the sample-wise quality and filter out noisy samples(e.g. the case in Fig.~\ref{fig:abstarct}(b)). Moreover, each component of GMM may be adapted to a semantic pattern of the instance (e.g. head, body of a person) in spatial dimensions, which may contain more semantic information and high-quality prediction. For each ground-truth $\mathcal{G}_i$, the probability density function of the quality distribution is formulated as follows:

\begin{equation}
\begin{aligned}
p_i(\vec{d}|\mathcal{G}_i, I,\theta)
=\sum_{k=1}^{K} \vec{\pi}_{i, k} \Phi_{i, k}(\vec{d}|\mathcal{G}_i, I,\theta),
% & = \sum_{k=1}^{K} \vec{\pi}_{k} \int p(\vec{x}|\mathcal{G}_i,z;I,\theta)p(z)dz
\end{aligned}
\label{equation:distribution}
\end{equation}
where the parameters $K, \theta$ are component number, parameters of the encoder. $\vec{\pi}$ denotes mixing coefficient along x- and y-axis in the spatial dimension of image $I$, and $\vec{d}$ represents the offsets of the sample inside an object to the GT's center along x- and y-axis. Moveover, we normalize $\vec{d}$ into range [-1, 1] by the ground-truth size. Here, we define each component of the 
mixture model as:
$
\Phi_{k}(\vec{d})
=\mathcal{N}_{k}(\vec{d}|\vec{\mu}_{k}, \vec{\sigma}_{k})
=e^{-\frac{(\vec{d}-\vec{\mu}_{k})^2}{2\vec{\sigma}_{k}^2}}.
$

\begin{figure}[t]
\centering 
\includegraphics[height=0.2\textwidth]{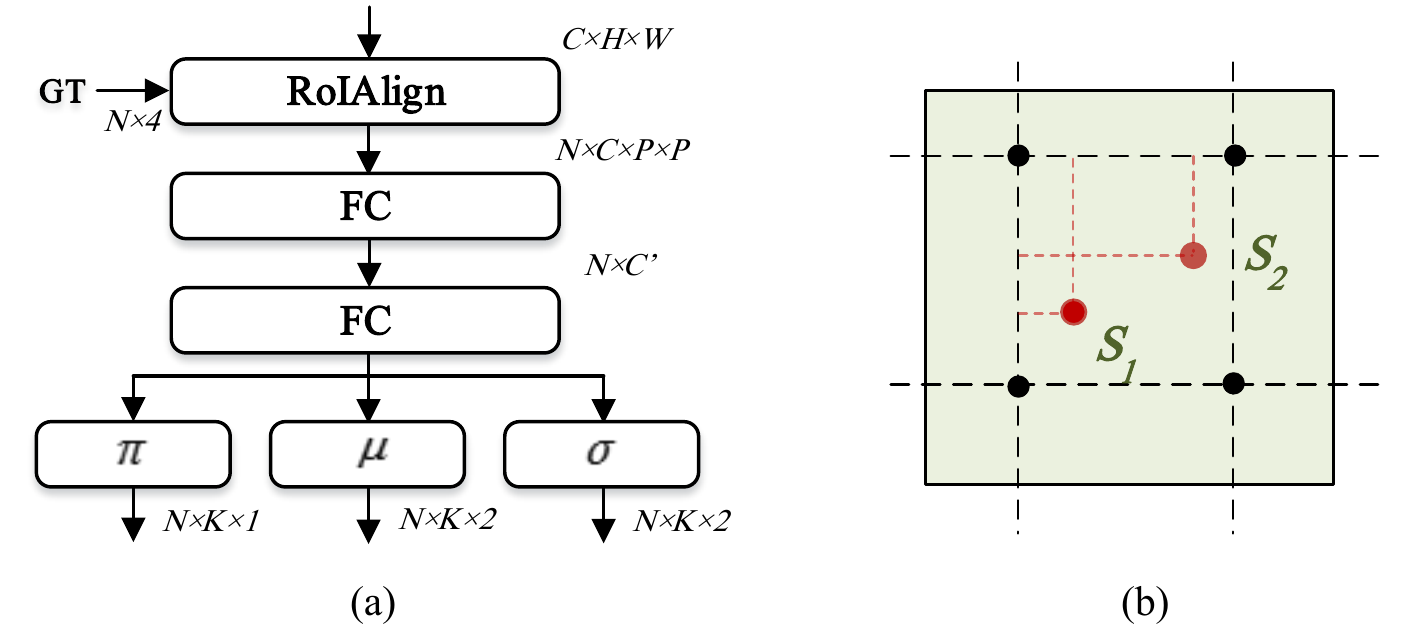}
\caption{(a) The details of the quality distribution encoder. $N,K$ represent the number of GTs, components of distributions. $C,H,W$ are the feature pyramid sizes. $\mu, \sigma$ denote location, scale of the distribution along x-axis and y-axis, and $\pi$ is the mixing coefficient. (b) We use bilinear interpolation~\cite{bilinear} to compute the exact values of the prediction. Dotted lines are the regular grid of feature map, and green square represent ground-truth $\mathcal{G}$. The red points are the sampling locations of $\mathcal{G}$. } 
\label{fig:encoder_and_bilinear}
\end{figure}

The network architecture of the encoder is shown in Fig.~\ref{fig:encoder_and_bilinear}(a). For simplicity, the encoder consists of two fully-connected layers, followed by fully-connected layers to predict the learnable parameters. The outputs of learnable encoder are  $\vec{\mu}_{n,k}, \vec{\sigma}_{n,k} \in \mathbb{R}^{N \times K \times 2}$ along spatial dimensions and $\vec{\pi}_{n,k} \in \mathbb{R}^{N \times K \times 1}$, which represent the location, scale, and mixing coefficient of the ground-truth $\mathcal{G}_n$' k-th component respectively. Specifically, the encoder share parameters $\theta$ among feature pyramid, and the distributions are estimated separately without feature fusion. To simplify our pipeline, we directly use the latent representation as the quality distribution. 

% \begin{equation}
% \begin{aligned}
% p_{qd}(\vec{d}|\vec{\mu_{i}}, \vec{\sigma_{i}}, \vec{\pi_{i}})
% =\sum_{k=1}^{K}\pi_{i, k}\Phi (\vec{d}|\vec{\mu}, \vec{\sigma}))
% =\sum_{k=1}^{K}\pi_{i, k}e^{-\frac{(\vec{d}-\vec{\mu})^2}{2\vec{\sigma}^2}}
% \end{aligned}
% \end{equation}

According to the QDE, our goal is to encode the ground-truth feature into quality distribution, which guides the sampling to select more high-quality samples. Thus we supervise the distribution by sample-wise quality metric (e.g. IoU of the predicted boxes). On the one hand, the quality distribution should be learnable and supervised by the quality of each prediction. QDE is a fully differentiable subnet which extracts the GT feature and jointly estimates the overall distribution of an instance. On the other hand, the distribution approximates the sample-wise quality by using GMM, which is easy to learn and noise-robust. Thus, we supervise the quality distribution by the IoU of the predicted boxes over the grid. 

\subsection{Quality Distribution for Sampling}
Based on the quality distribution, we formulate the sampling and label assignment strategy in IQDet. First, instead of sampling over the grid, we resample floating-number coordinate predictions from the distribution. Meanwhile, the distribution represents the quality of these predictions, thus we supervise their scores with the soft distribution.

% \begin{figure}[t]
% \centering 
% \includegraphics[height=0.15\textwidth]{images/bilinear.pdf}
% \label{fig:bilinear}
% \caption{We use bilinear interpolation\cite{bilinear} to compute the exact values of the prediction. Dotted lines are the regular grid of feature map, and green square represent ground-truth $\mathcal{G}$. The red points are the sampling locations of $\mathcal{G}$. We use interpolation\cite{bilinear} to sample the prediction of the floating-number locations. }
% \end{figure}

% \vspace{0pt}
\paragraph{Resample in Spatial and Scale Dimensions.}
\label{sec:quality_sampling}
Previous one-stage detector sample over regular grids of feature map, and they usually select \textit{topK} or IoU thresholds to reject close false positives. However, forcing larger IoU thresholds or smaller \textit{topK} leads to exponentially smaller numbers of positive training samples. They may get involved in the problems of overfitting at training when the ground-truths are assigned with insufficient positive samples. 

Based on the distributions mentioned in Sec.~\ref{sec:quality_distribution_encoder}, our sampling strategy select positive samples in a probabilistic manner, and a more diverse and high-quality positive training set is available. First, based on the distribution probability, we resample a fixed number of predictions over floating-number locations, namely sampling continuous coordinates over the feature map. The red circles and orange circles in QDS in Fig.~\ref{fig:architecture} represent the positive samples to be trained. The predictions over these floating-number locations are hard to obtain. Thus, we avoid any quantization of these prediction's locations by using bilinear interpolation~\cite{bilinear} to compute the exact values of the predictions. The interpolation operators are calculated over the classification and regression prediction maps. Under the sampling strategy, IQDet is more likely to resample the predictions which are close to the center of the Gaussian. Therefore, we can resample sufficient and high-quality samples instead of sampling over regular grids of feature maps.

According to the bilinear interpolation of the predictions, we construct a bag of candidate predictions for each object $\mathcal{G}$ by resampling \textit{K} positive predictions in terms of their quality distribution values. Based on the spatial quality sampling above, we use the distribution to select the training positive samples in a probabilistic manner. For the negative samples, our sampling strategy is similar to the baseline. Namely, we select the negative sample outside the GT box over the grid. It should be noted that our sampling strategy is different from the loss reweighting strategy. As shown in ~\ref{fig:encoder_and_bilinear}(b), owing to the classification feature map is a nonlinear distribution, thus the IQDet back-propagates different gradient to the grid point. Moreover, the interpolation for each sample is similar to the RoIAlign~\cite{mask_rcnn} when the \textit{poolsize} equals to 1. It is well aligned and preserves the per-pixel spatial correspondence. To sum up, compared with the previous strategies sampling over the grid, IQDet is able to train with more diverse and high-quality samples.

\paragraph{Soft Label Assignment.}
For each sample selected from the quality distribution, we also calculate the label according to bilinear interpolation. Different from the label assignment in previous work, we use the quality distribution as the labels to supervise the classification prediction. Without dividing the samples into positive and negative samples, IQDet assigns the quality distribution as a soft label to each sample. The predictions with high-quality value are more likely supervised by higher classification target. For predictions located outside the ground-truth region, we assign them as negative samples for sure, which the classification targets are set to zero. It is worth noting that Quality Distribution Sampling (QDS) is applied to all feature pyramid levels. Due to the GMM variables of an instance in each FPN layer are different, the distributions encoded by different pyramid layers are not similar. Namely, for an instance in different pyramid layers, we assign the label with different quality distributions independently. Across all feature pyramid levels, only the \textit{K} samples in terms of distribution values are supervised by the distribution, and the others are supervised as zero. For the regression branch, we use the offset of the sampling location with ground-truth as the regression target, which is similar to the previous works. Furthermore, to compensate for the interference caused by the different downsampling rates of FPN, we normalize the distance $\vec{d}$ by its FPN stage’s downscale ratio.

\subsection{IQDet}
\paragraph{Network Architecture.}
We now present the network architecture of our IQDet. The framework of IQDet is demonstrated in Fig.~\ref{fig:architecture}. In our experiments, we adopt a simple anchor-free object detector FCOS~\cite{fcos} as our baseline. The network consists of a backbone, a feature pyramid, and one detection head per pyramid level, in a fully convolutional network. The pyramid levels are denoted as $P_3-P_7$, which is similar to the baseline. 

The Quality Distribution Subnet is shown in the orange box of Fig.~\ref{fig:architecture}. First, taking the pyramid feature maps as input, the Quality Distribution Encoder extracts the regional feature of the ground-truth, then it estimates a quality distribution which guides the sampling and label assignment strategy. The operators in this Quality Distribution Subnet are instance-wise, and the samples of an object are estimated and supervised jointly. Moreover, This subnet acts as an auxiliary supervision to the detection backbone during training and it is completely cost-free in inference time. In Class+Box Subnet, the operators are sample-wise, and each sample is estimated and supervised separately. The subnet architecture is exactly the same as the one in FCOS~\cite{fcos} and ATSS~\cite{atss}, which is RetinaNet~\cite{focal_loss} with modified feature subnets and an auxiliary prediction head. In our experiments, we use predicted IoUs with corresponding GT boxes, instead of centerness scores when using the auxiliary prediction as in ~\cite{fcos}.

\paragraph{Training and Inference.}
The proposed IQDet is easy to optimize in an end-to-end way using a multi-task loss. Combining the output of the IQDet, we define our training loss function as follows:

\begin{equation}
\begin{aligned}
\begin{split}
\mathcal{L}= 
& \mathcal{L}_{cls} + \mathcal{L}_{reg} + \mathcal{L}_{aux} + \lambda_{IQ} \mathcal{L}_{IQ} \\
\end{split}
\end{aligned}
\end{equation}

In the implementation, focal loss~\cite{focal_loss}, IoU loss and Binary Cross-Entropy (BCE) loss are used as the classification loss, regression loss and auxiliary loss respectively, which are the same as FCOS~\cite{fcos}. For the $\mathcal{L}_{IQ}$ loss shown in Fig.~\ref{fig:architecture}, we supervise the quality distribution by the IoUs of predicted boxes. As the loss function, we use the BCE loss between the target values and the distribution predictions.

% #################################################################################
%                              Experiments
% #################################################################################
\section{Experiments}
\subsection{Implementation Details}
Following the common practice, our experiments are trained on the large-scale detection benchmark COCO trainval35k set (115K images) and evaluated on COCO val set (5K images). To compare with the state-of-art approaches, we report COCO AP on the test-dev set (20K images).
\paragraph{Training Details.}
We use ResNet-50 with FPN as our backbone network for all the experiments, if not specified. We use synchronized stochastic gradient descent (SGD) over 8 GPUs with a total of 16 images per minibatch (2 images per GPU) for 90k iterations. With an initial learning rate of 0.01, we decrease it by a factor of 10 after 60k iterations and 80k iterations respectively. Weight decay of 0.0001 and momentum of 0.9 are used. 
% We initialize our backbone network with the weights pre-trained on ImageNet. We use horizontal image flipping as the only form of data augmentation. 
And the input images are resized to ensure their shorter edge being 800 and the longer edge less than 1333 unless noted. In our experiments, $\lambda_{IQ}$ is 1 and the number of GMM components equals 2, and our method is not sensitive to these values. 
\paragraph{Inference Details.}
During the inference phase, we resize the input image in the same way as in the training phase, and then forward it to obtain the predicted bounding boxes with a predicted class. The following post-processing is exactly the same with RetinaNet and we directly use the same post-processing hyper-parameters (such as the threshold of NMS) of RetinaNet. Finally, the Non-Maximum Suppression (NMS) is applied with the IoU threshold 0.6 per class to yield the final top 100 confident detections per image.

\subsection{Ablation Study}

\begin{table}[t]
\setlength{\tabcolsep}{4pt}
\centering
\caption{Ablation study of Quality Distribution Encoder (QDE) and Quality Distribution sampling (QDS). ``$\ast$'' denotes centerness as auxiliary task, while others use IoU prediction.}
\resizebox{0.34\textwidth}{42pt}{
\begin{tabular}{c|cc|c|cc}
\toprule
Method & QDE        & QDS  & $AP$ &$AP_{50}$&$AP_{75}$\\
\midrule
FCOS$\ast$ &     &            & 38.7 &  57.5   &  41.7 \\
FCOS  &     &            & 39.4 &  57.9   &  42.4 \\
\midrule
IQDet & \checkmark &            & 40.4 &  57.7   &  44.1   \\
IQDet &            & \checkmark & 39.6 &  57.5   &  43.0   \\
IQDet & \checkmark & \checkmark & \textbf{41.1} &\textbf{58.9}&\textbf{44.8} \\
\bottomrule
\end{tabular}
}
\label{tab:ablation_study}
\end{table}

\subsubsection{Quality Encoder and Quality Sampling.}
To demonstrate the effectiveness of the two key components, we gradually add the Quality Distribution Encoder (QDE) and Quality Distribution sampling (QDS) to the baseline to investigate the performance of our proposed IQDet. Based on the FCOS as our baseline, we first analyze the impact of QDE. According to the quality distribution, we just use the soft label assignment and sample over the grid of the feature map. As the results in the second row of Table.~\ref{tab:ablation_study}, QDE brings relatively significant performance gain by 1.7 AP, which suggests that the quality distribution is critical for guiding the training. Then we also analyze the effectiveness of QDS in the third row of the table, where the sampling distribution is only based on the center region. We select positive samples with continuous coordinates. The proposed QDS also improves FOCS by 0.9 AP. Finally, the implementation of the IQDet can improve the performance by 2.4 AP in total.

%=====================  EXperiments of Quality Distribition ===================== 
\subsubsection{Quality Distribution Encoder.}
\paragraph{Learnable Parameters of Quality Distribution.} 
IQDet uses Gaussian Mixture Model (GMM) to estimate the quality distribution, which guides the sampling strategy. To analyze the design of the quality distribution, we compare different quality distributions in Table.~\ref{tab:gaussian_param}. To simplify the analysis, we denote the``\checkmark'' as learnable parameters and analyze the learnable part of the mixture distribution to verify the gains of each part. We can see that the learnable $\vec{\mu}$ leads to a gain of nearly 10 AP and the learnable $\vec{\sigma}$ can further improve 1.4 AP. Learnable peak $\vec{p}$ of the gaussian distribution shows improvements of 0.4 AP. From the whole table, we can see that the quality distribution brings major improvement and the design of learnable quality distribution is critical.
\begin{table}[t]
\caption{The ablation experiments of learnable parameters in quality distribution. ``\checkmark'' means this parameter is learnable.}
\centering
\resizebox{0.44\textwidth}{42pt}{
\begin{tabular}{c|ccc|ccccccc}
\toprule
& $\vec{\mu}$ & $\vec{\sigma}$ & $\vec{p}$ &  $AP$  & $AP_{50}$ & $AP_{75}$ \\
\midrule
Center Region &            &            &  & 38.7 & 57.5 & 41.7   \\
\midrule
~              &            &            &  & 29.8 & 47.9 & 32.1   \\
IQ-Distribution& \checkmark &            &  & 38.6 & 57.2 & 41.8   \\
(GMM)     & \checkmark & \checkmark &  & 40.0 & 57.9 & 43.3                                  \\
        ~ & \checkmark & \checkmark & \checkmark & \textbf{40.4} & \textbf{58.4} & \textbf{43.6} \\
\bottomrule
\end{tabular}
}
\label{tab:gaussian_param}
\end{table}

\paragraph{Architecture of Encoder.} 
To analyze the impact of the RoI feature extractor, we compare the different architecture of feature extractor, shown in Table.~\ref{tab:encoder}. We notice that RoiAlign extractor achieves 41.1 AP in the third row, outperforming the counterpart with pooling by 1.0 AP. It demonstrates that the spatial information is crucial for our distribution modeling. We also compare different pooling size of RoIAlign. Considering the speed/accuracy trade-off, pooling size equals 7 in all our experiments.

\begin{table}[t]
\setlength{\tabcolsep}{4pt}
\centering
\caption{For the second row, we use the RoIAlign to extract feature, then we eliminate the spatial information by a pooling layer. We also compare different pooling size of RoIAlign.}
\resizebox{0.44\textwidth}{46pt}{
\begin{tabular}{c|c|cccccc}
\toprule
Methods  & $AP$ & $AP_{50}$ & $AP_{75}$ & $AP_{S}$& $AP_{M}$ & $AP_{L}$\\
\midrule
RoIPool~\cite{fasterrcnn}   & 40.6 & 58.3 & 44.0 & 23.5 & 44.4 & 53.2 \\
RoIAlign (pool)             & 40.1 & 58.3 & 43.6 & 23.0 & 44.0 & 52.5 \\
RoIAlign~\cite{mask_rcnn}   & \textbf{41.1} & \textbf{58.9} & \textbf{44.8} & \textbf{23.5} & \textbf{44.9} & 54.4 \\
\midrule
RoI ($3\times3$)     & 40.6 & 58.8 & 44.4 & 23.2 & 44.5 & 53.7 \\
RoI ($7\times7$)     & \textbf{41.1} & \textbf{58.9} & 44.8 & 23.5 & \textbf{44.9} & 54.4 \\
RoI ($14\times14$)   & \textbf{41.1} & 58.8 & \textbf{44.9} & \textbf{23.7} & \textbf{44.9} & \textbf{54.6} \\
\bottomrule
\end{tabular}
}
\label{tab:encoder}
\end{table}

%=====================  EXperiments of Quality sampling ===================== 
\subsubsection{Quality Distribution Sampling.}

\begin{figure*}[htbp]
\centering
\includegraphics[height=0.25\linewidth, width=0.9\linewidth]{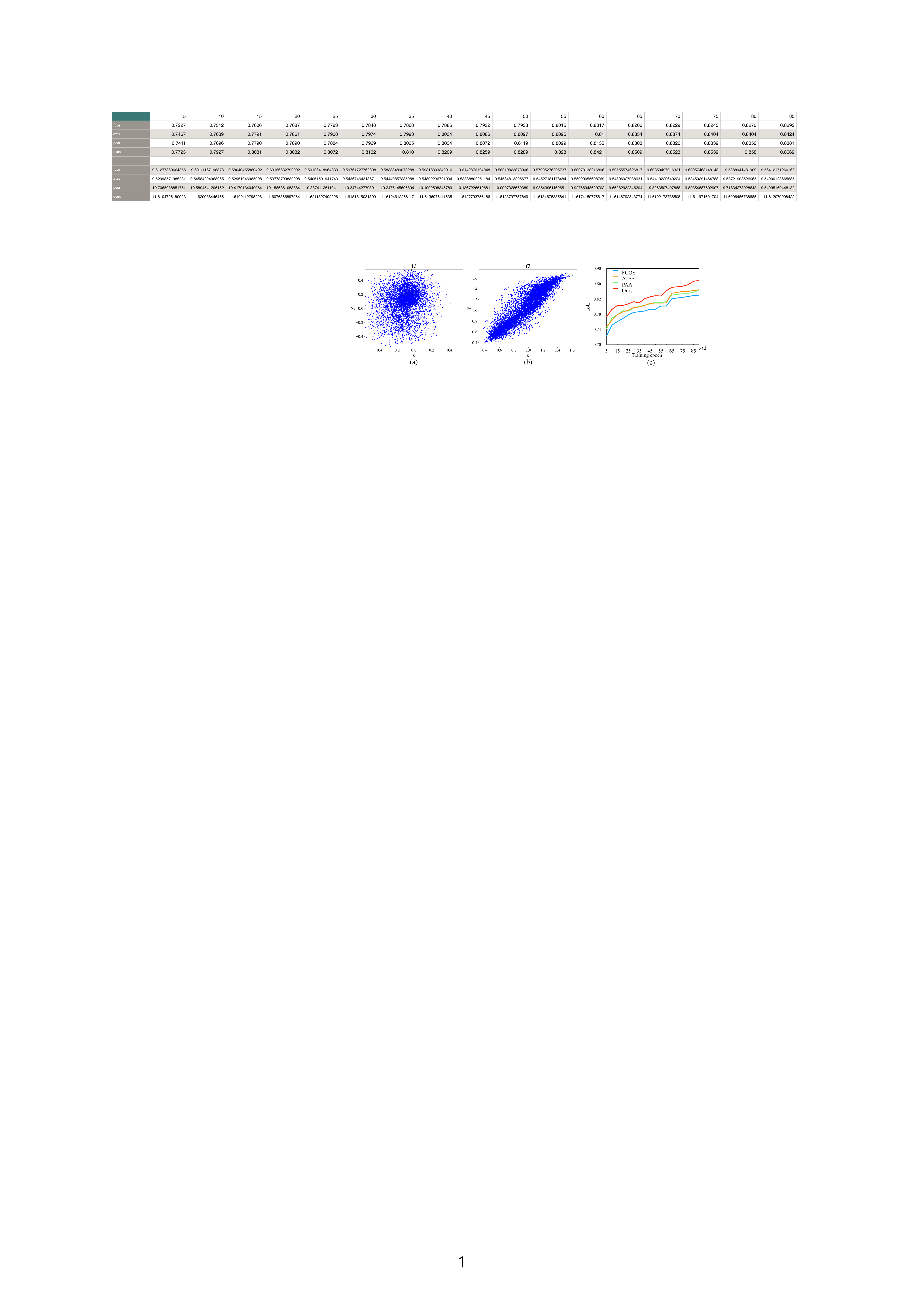}
\label{fig:analysis}
\caption{(a)(b) are the $\mu, \sigma$ of Gaussian components respectively and each blue point denotes a quality distribution for an instance. (c) Comparison of IoU curves for different sampling strategies, it suggests that IQDet trains with more high-quality samples. }
\end{figure*}

\paragraph{Quantified Sampling vs. Continuous Sampling}
As described in Sec.~\ref{sec:quality_distribution_encoder}, IQDet selects samples according to quality distribution. We conduct several experiments to study the effect of the quality sampling strategy. In our experiment, we first select the samples by uniform distribution in spatial dimensions, which is shown in the first row of Table.~\ref{tab:sampling}. Then we use quality distribution to sample, but we quantify the locations by rounding off the coordinates. The quality distribution sampling with quantified location show 0.4 AP relative improvement over the baseline. Finally, we use bilinear interpolation to supervise the train samples with continuous coordinates. Specifically, IQDet uses the value of the distribution as the probability to sample, leading to sample more high quality and sufficient samples than other sampling strategies. It increases the performance by 0.7 AP which demonstrating that sampling sufficient and high-quality bounding boxes by QDS will improve detector performance. 

\begin{table}[t]
\setlength{\tabcolsep}{4pt}
\centering
\caption{Analysis of Quality sampling. ``\dag'' represents that we select the prediction with bilinear interpolation~\cite{bilinear}, while the experiments without ``\dag'' use quantified locations to sample the prediction to train.}
\resizebox{0.42\textwidth}{30pt}{
\begin{tabular}{c|c|cccccc}
\toprule
Methods  & $AP$ & $AP_{50}$ & $AP_{75}$ & $AP_{S}$& $AP_{M}$ & $AP_{L}$\\
\midrule
Uniform      & 40.4 & 57.7 & 44.1 & 22.7 & 44.1 & 53.1 \\
QD           & 40.8 & 58.6 & 44.6 & 23.4 & 44.4 & 54.1 \\
QD\dag(ours) & \textbf{41.1} & \textbf{58.9} & \textbf{44.8} & \textbf{23.5} & \textbf{44.9} & \textbf{54.4} \\
\bottomrule
\end{tabular}
}
\label{tab:sampling}
\end{table}

\paragraph{Number of Positive Samples of Resampling.}
Moreover, IQDet selects \textit{K} positive samples in terms of their quality distribution values. It achieves better performance when the number of positive samples is set to 12, shown in Table.~\ref{tab:num_resample}.

\begin{table}[t]
\centering
\caption{Varying \textit{K} for the number of positive predictions resampling based on quality distribution.}
\resizebox{0.44\textwidth}{40pt}{
\begin{tabular}{c|c|cc|ccc}
\toprule
\textit{K} & $AP$ & $AP_{50}$ & $AP_{75}$ & $AP_{S}$& $AP_{M}$ & $AP_{L}$\\
\midrule
4  & 40.6 & 58.4 & 44.0 & 23.4 & 44.4 & 53.8 \\
8  & 40.6 & 58.6 & 44.1 & 23.1 & 44.1 & 53.6 \\
12 & \textbf{41.1} & \textbf{58.9} & \textbf{44.8} & 23.5 & \textbf{44.9} & \textbf{54.4} \\
16 & 40.6 & 58.3 & 44.5 & 23.4 & 44.5 & 53.0 \\
20 & 40.5 & 58.4 & 44.0 & \textbf{23.6} & 44.3 & 53.5\\
\bottomrule
\end{tabular}
}
\label{tab:num_resample}
\end{table}

\paragraph{Quality Distribution Target.}
We conduct additional ablation studies regarding the target of quality distribution. The quality of the bounding box can be represented in many forms (e.g. IoU or training loss) to guide the label assign. We compare the performance of detectors with different forms of quality distribution targets. In the first three rows, the predicted boxes with the \textit{K} smallest $L_{cls}$ and $L_{reg}$ as the positives samples, and we normalize the loss value into the range [-1, 1]. Table.~\ref{tab:quality_target} shows that loss-based label assignments are comparable. For the comb-loss, we combine the $L_{cls}$ and $L_{reg}$ to guide the label assign, marked as \textit{comb-loss} in the Table. Finally, the last row of the table is our label assign strategy which uses the IoU of predicted boxes to guide the label assign strategy. We also verify that IoU prediction is more effective than the other forms of the assignments (40.6 AP vs. 41.1 AP).

\begin{table}[t]
\centering
\setlength{\tabcolsep}{4pt}
\caption{Ablation study of quality distribution target. ``Comb-loss'' denotes the combined loss of the classification loss and regression loss.}
\resizebox{0.44\textwidth}{40pt}{
\begin{tabular}{c|c|cc|ccc}
\toprule
~         & AP & $AP_{50}$ & $AP_{75}$ & $AP_{S}$& $AP_{M}$ & $AP_{L}$\\
\midrule
Cls-loss  & 40.7 & 58.4 & 44.4 & \textbf{24.2} & 44.1 & 53.3 \\
Reg-loss  & 40.6 & 58.7 & 43.9 & 23.1 & 44.6 & 54.0 \\
Comb-loss & 40.6 & 58.2 & 44.2 & 23.9 & 44.3 & 54.0 \\
\midrule
Ours(IoU)  & \textbf{41.1} & \textbf{58.9} & \textbf{44.8} & 23.5 & \textbf{44.9} & \textbf{54.4}\\
\bottomrule
\end{tabular}
}
\label{tab:quality_target}
\end{table}

\begin{table*}[!t]
\centering
\caption{IQDet vs. the state-of-the-art mothods (single model) on COCO test-dev set. ``$\dag$'' indicates the multi-scale testing}
\resizebox{0.86\textwidth}{130pt}{
\begin{tabular}{l|c|c|cccccc}
\toprule
Method                       & Iteration & Backbone & AP & $AP_{50}$& $AP_{75}$ & $AP_{S}$ & $AP_{M}$ & $AP_{L}$ \\
\midrule
FCOS~\cite{fcos}             & 180k  & ResNet-101-FPN & 41.5 & 60.7 & 45.0 & 24.4  & 44.8 & 51.6  \\
FreeAnchor~\cite{freeanchor}  & 180k  & ResNet-101-FPN & 43.1 & 62.2 & 46.4 & 24.5  & 46.1 & 54.8  \\
SAPD~\cite{sapd}              & 180k  & ResNet-101-FPN & 43.5 & 63.6 & 46.5 & 24.9  & 46.8 & 54.6  \\
ATSS~\cite{atss}              & 180k  & ResNet-101-FPN & 43.6 & 62.1 & 47.4 & 26.1  & 47.0 & 53.6  \\
PAA~\cite{paa}                & 180k  & ResNet-101-FPN & 44.6 & 63.3 & 48.4 & 26.4  & 48.5 & 56.0  \\
IQDet(Ours)                  & 180k  & ResNet-101-FPN & \textbf{45.1} & \textbf{63.4} & \textbf{49.3} & \textbf{26.7}  & \textbf{48.5} & \textbf{56.6}\\
\midrule
FCOS~\cite{fcos}             & 180k  & ResNeXt-64x4d-101 & 44.7 & 64.1 & 48.4 & 27.6 & 47.5 & 55.6  \\
FreeAnchor~\cite{freeanchor}  & 180k  & ResNeXt-64x4d-101 & 44.9 & 64.3 & 48.5 & 26.8 & 48.3 & 55.9  \\
SAPD~\cite{sapd}              & 180k  & ResNeXt-64x4d-101 & 45.4 & 65.6 & 48.9 & 27.3 & 48.7 & 56.8  \\
ATSS~\cite{atss}              & 180k  & ResNeXt-64x4d-101 & 45.6 & 64.6 & 49.7 & 28.5 & 48.9 & 55.6  \\
PAA~\cite{paa}                & 180k  & ResNeXt-64x4d-101 & 46.3 & 65.3 & 50.4 & 28.6 & 50.0 & 57.1  \\
IQDet(Ours)                  & 180k  & ResNeXt-64x4d-101 & \textbf{47.0} & \textbf{65.7} & \textbf{51.1} & \textbf{29.1} & \textbf{50.5} & \textbf{57.9} \\
\midrule
SAPD~\cite{sapd}              & 180k  & ResNeXt-64x4d-101-DCN & 45.4 & 65.6 & 48.9 & 27.3 & 48.7 & 56.8  \\
ATSS~\cite{atss}              & 180k  & ResNeXt-64x4d-101-DCN & 47.7 & 66.5 & 51.9 & 29.7 & 50.8 & 59.4 \\
PAA~\cite{paa}                & 180k  & ResNeXt-64x4d-101-DCN & 48.6 & 67.5 & 52.7 & 29.9 & 52.2 & 61.5 \\
IQDet(Ours)                  & 180k  & ResNeXt-64x4d-101-DCN & 49.0 & 67.5 & 53.1 & 30.0 & 52.3 & 62.0 \\
\midrule
ATSS~\cite{atss}$\dag$        & 180k  & ResNeXt-64x4d-101-DCN & 50.7 & 68.9 & 56.3 & 33.2 & 52.9 & 62.2 \\
PAA~\cite{paa}$\dag$          & 180k  & ResNeXt-64x4d-101-DCN & 51.3 & 68.8 & 56.6 & 34.3 & 53.5 & 63.6 \\
IQDet(Ours)$\dag$            & 180k  & ResNeXt-64x4d-101-DCN & 51.6 & 68.7 & 57.0 & 34.5 & 53.6 & 64.5 \\
\bottomrule
\end{tabular}
}
\label{tab:sota}
\end{table*}

%=====================  Analysis of IQDet ===================== 
\begin{figure*}[htb]
\centering
\includegraphics[height=0.27\linewidth, width=0.9\linewidth]{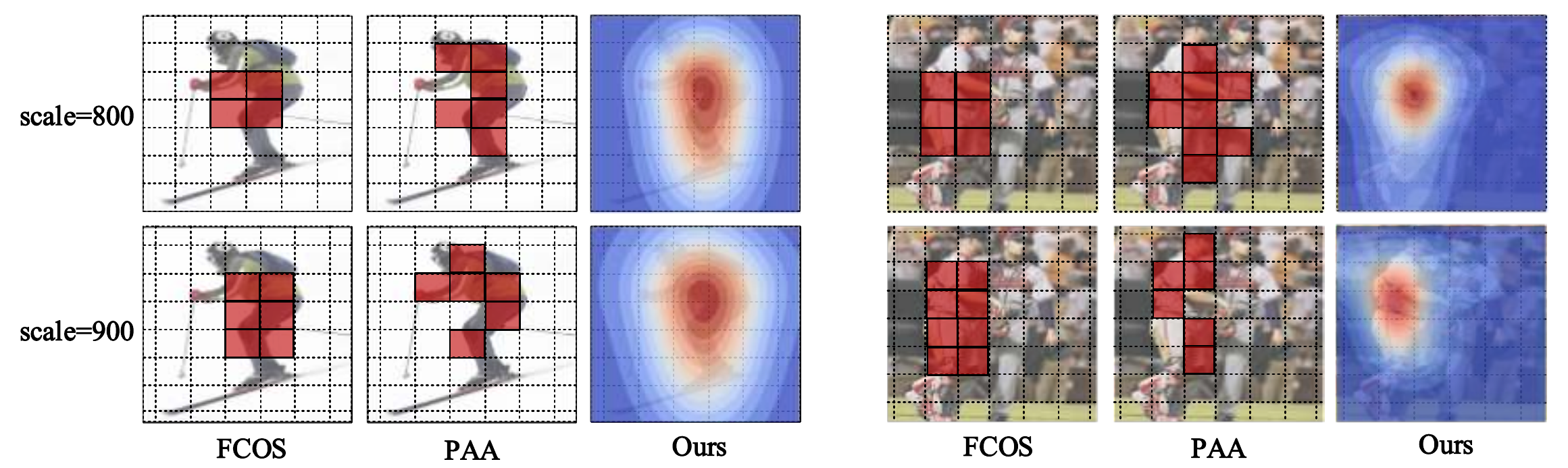}
\caption{The visualization of label assignment and sampling. The red points are the sampling locations of these strategies.}
\label{fig:latent_vis}
\end{figure*}

\subsection{Analysis of IQDet}

\paragraph{Analysis of Quality Distribution.}
We evidence our experiments by visualization of distribution representation. We can see that the peaks of the distribution are mainly distributed in some semantic parts of an object, such as the head and trunk of the human in Fig.~\ref{fig:latent_vis}. For the sampling locations, IQDet is able to select more samples with higher quality than the other strategies.
We also provide some statistics on the $\mu, \sigma$ of quality distribution. As shown in Fig.~\ref{fig:analysis}(a)(b), each blue point in this figure represents the value of $\mu$ or $\sigma$ for the distribution. The statistics show that sampling over the center region is not always the best choice for an instance in COCO. 

\paragraph{Training curves of Quality Distribution Sampling.}
The train curves of different strategies are depicted in Fig.~\ref{fig:analysis}(c). Obviously, the average IoUs of positive samples' in IQDet are higher than the other methods, and it demonstrates that training with more high-quality samples can boost the performance of the baseline.

%=====================  Comparison with State-of-the-art Detectors ===================== 
\subsection{Comparison with State-of-the-art Detectors.}
We compare IQDet with other state-of-the-art detectors on MS COCO test-dev set. We adopt 180K training iterations following the previous works. Results are shown in Table.~\ref{tab:sota}. Our IQDet with ResNet-101 backbone achieves 45.1 AP, outperforming all state-of-the-art one-stage detectors with the same backbone as we known. By changing the backbone and training setting the same to other methods, our method consistently surpasses the label assign approaches including ATSS, PAA. By adopting advanced settings, IQDet reaches 51.6 AP, the state of the art among existing one-stage cost-free methods.

%=====================  Generalization of IQDet ===================== 
\subsection{Generalization of IQDet.}
To prove the generalization ability of the IQDet, we evaluate our proposed methods with other cost-free one-stage detectors on different detection datasets. For a fair comparison, without modifying any setting of these methods, we only adjust the training setting following the common paradigm of each dataset. As shown in Table.~\ref{tab:generalization}, IQDet still has a strong generalization ability on other datasets. Especially for the WiderFace dataset, our method has 1.4 AP gain than all state-of-the-art one-stage detectors on WiderFace Dataset.

\begin{table}[t]
\caption{Generalization ability of IQDet. Performance comparison with typical one-stage detectors on PASCAL VOC and WiderFace.}
\resizebox{0.50\textwidth}{40pt}{
\centering
\begin{tabular}{c|ccc|ccc}
\toprule
\multirow{2}*{Method} & \multicolumn{3}{c|}{PASCAL VOC~\cite{pascal_voc}} & \multicolumn{3}{c}{WiderFace~\cite{widerface}} \\
\cline{2-7}
~                       & AP & $AP_{50}$ & $AP_{75}$ & AP & $AP_{50}$ & $AP_{75}$ \\
\midrule
FCOS$^{\ast}$~\cite{fcos}         & 56.2 & 79.9 & 61.9 & 50.6 & 89.1 & 51.7 \\
ATSS~\cite{atss}          & 56.7 & 79.9 & 62.3 & 51.6 & 89.5 & 53.9 \\
PAA(w/o vote)~\cite{paa}  & 58.3 & 80.7 & 64.4 & 51.0 & 87.2 & 53.8 \\
IQDet(Ours)              & \textbf{59.0} & \textbf{81.4} & \textbf{65.0} & \textbf{52.4} & \textbf{89.6} & \textbf{55.1} \\
\bottomrule
\end{tabular}
}
\label{tab:generalization}
\end{table}

\section{Conclusion}

In this paper, we proposed an instance-wise quality distribution sampling framework, called IQDet. It is a learnable and prediction-aware sampling strategy according to a mixture model. Our sample selection is guided by a quality distribution which automatically selects samples according to the spatial pattern of the object. This architecture was shown to select more high-quality predictions and avoid the problems of overfitting at training. This method achieves nearly 2.5 AP boost over baseline, and surpassed all previous one-stage object detector on COCO test-dev set. Besides, extensive experiments on other datasets demonstrate that IQDet can conveniently transfer to other datasets and tasks without additional modification. We believe that it can be useful for many future object detection research efforts.

\section*{Acknowledgement}
This research was partially supported by National Key RD Program of China (No. 2017YFA0700800), and Beijing Academy of Artificial Intelligence (BAAI).

\newpage

{\small
\bibliographystyle{ieee_fullname}
\bibliography{egbib}
}

\end{document}